\documentclass[runningheads]{llncs}
\usepackage[T1]{fontenc}
\usepackage{graphicx}
\usepackage{ragged2e} 
\usepackage{booktabs,makecell, multirow, tabularx}
\usepackage[backref]{hyperref}
\usepackage{array} 
\usepackage{arydshln}
\usepackage[cmyk]{xcolor}
\usepackage{amsmath,bm}
\usepackage{amssymb}

\begin{document}
\title{Few-Shot Multi-Label Aspect Category Detection Utilizing Prototypical Network with Sentence-Level Weighting and Label Augmentation}
\titlerunning{Few-Shot Multi-Label ACD Utilizing PN w/ SLW and LA}
%
\author{Zeyu Wang \orcidID{0009-0003-7889-7583}\and
Mizuho Iwaihara\thanks{Corresponding author}\orcidID{0000-0001-6985-9671}}
\authorrunning{Wang et al.}
%
\institute{Graduate School of Information, Production,and  Systems,\\ Waseda University, Kitakyushu 808-0135, Japan\\
\email{wangzeyu@akane.waseda.jp},
\email{iwaihara@waseda.jp}\\}
\maketitle              
\begin{abstract}
Multi-label aspect category detection is intended to detect multiple aspect categories occurring in a given sentence. Since aspect category detection often suffers from  limited datasets and data sparsity, the prototypical network with attention mechanisms has been applied for few-shot aspect category detection. Nevertheless, most of the prototypical networks used so far calculate the prototypes by taking the mean value of all the instances in the support set. This seems to ignore the variations between instances in multi-label aspect category detection. Also, several related works utilize label text information to enhance the attention mechanism. However, the label text information is often short and limited, and  not specific enough to discern categories. In this paper, we first introduce support set attention along with the augmented label information to mitigate the noise at word-level for each support set instance. Moreover, we use a sentence-level attention mechanism that gives different weights to each instance in the support set in order to compute prototypes by weighted averaging. Finally, the calculated prototypes are further used in conjunction with query instances to compute query attention and thereby eliminate noises from the query set. Experimental results on the Yelp dataset show that our proposed method is useful and outperforms all baselines in four different scenarios.

\end{abstract}
\section{Introduction}

\begin{table}[t]
	\caption{Example of a 3-way 2-shot meta-task. The bolded parts with gray background represent the target aspects, while the square marked parts indicate the noise aspects.}\label{tab1}
	
	\resizebox{\linewidth}{!}{
		\begin{tabular}{c|c}
			\hline
			\multicolumn{2}{c} {\begin{slshape}Support Set\end{slshape}}\\
			\hline
			\multirow{2}{*}{\begin{slshape}(A) experience\end{slshape}} 
			& \makecell[l]{\begin{slshape}(1) Perhaps we’ll try one more time and hope our \colorbox{lightgray}{\textbf{experience}} is better.\end{slshape}} \\ \cdashline{2-2}[1pt/1pt] & \makecell[l]{\begin{slshape}(2) The \colorbox{lightgray}{\textbf{experience}} and \framebox{service} is very great!\end{slshape}}\\
			\hline
			
			\multirow{2}{*}{\begin{slshape}(B) drinks\end{slshape}} 
			& \makecell[l]{\begin{slshape}(1) It was happy hour so the \colorbox{lightgray}{\textbf{drinks}} were a little \framebox{less expensive}.\end{slshape}} \\ \cdashline{2-2}[1pt/1pt] & \makecell[l]{\begin{slshape}(2) Just an hour in the afternoon and only \framebox{50 cents} or so off the \colorbox{lightgray}{\textbf{drinks}} with no \framebox{food} specials.\end{slshape}}\\
			\hline
			
			\multirow{2}{*}{\begin{slshape}(C) food\end{slshape}} 
			& \makecell[l]{\begin{slshape}(1) They also have rotating \colorbox{lightgray}{\textbf{dining}} specials.\end{slshape}} \\ \cdashline{2-2}[1pt/1pt] & \makecell[l]{\begin{slshape}(2) The \colorbox{lightgray}{\textbf{food}} was good and \framebox{price} was reasonable.\end{slshape}}\\
			\hline
			
			\multicolumn{2}{c} {\begin{slshape}Query Set\end{slshape}}\\
			\hline
			
			{\begin{slshape}(A) and (C)\end{slshape}} 
			& \makecell[l]{\begin{slshape}My \colorbox{lightgray}{\textbf{experience}} as far as \framebox{service} and the \colorbox{lightgray}{\textbf{food}} are the same.\end{slshape}} \\
			\hdashline[1pt/1pt]
			
			{\begin{slshape}(B)\end{slshape}} 
			& \makecell[l]{\begin{slshape}\colorbox{lightgray}{\textbf{Drinks}} were tasty and quick, and the \framebox{atmosphere} was cool.\end{slshape}} \\
			
			\hline
	\end{tabular}}
\end{table}

Aspect category detection (ACD) \cite{pontiki2015semeval}\cite{pontiki2016semeval} is a sub-task of aspect-based sentiment analysis (ABSA) \cite{hu2004mining}. ACD is to categorize user reviews on products and services such as hotels and restaurants into a pre-defined set of aspect categories. Examples of aspect categories for hotels are location, price, room, while those of restaurants are food, service, interior, etc. ACD will facilitate access to viewpoint information for users and provide assistance for making decisions. As in practical scenarios, user reviews are generally diversified and contain more than one aspect, the task of multi-label aspect category detection becomes essential. This task can also be perceived as a special case of multi-label text classification tasks.

Few-shot learning (FSL) \cite{fe2003bayesian}\cite{fei2006one} enables a quick adaption to novel classes with a limited number of samples after learning a large amount of data, being an effective solution to the issues of finite data and data sparsity. FSL problems can be dealt with by a meta-learning \cite{hochreiter2001learning} strategy, which is also known as ``learning to learn.'' In the meta-training phase, the dataset is divided into separate meta-tasks to learn the generalization capability of the model in the case of category changes. The meta-task adopts the $ N $-way $ K $-shot setting, as demonstrated in Table \ref{tab1}, which is an example of a 3-way 2-shot meta-task, meaning that there are altogether three classes (aspect categories) in the support set and in each class there are two samples (sentences). The prototypical network \cite{snell2017prototypical} utilized in this paper follows exactly the meta-paradigm described above.

A prototypical network aims to extract a prototype for each class by averaging all the instances in one class to measure the distance with the instance of the query set. However, as shown in Table \ref{tab1}, it is evident that the number of noise aspects contained in the sentences of each support set class is different. In terms of that, simply averaging all instances in a class neglects the differences between sentences and treats samples from the same classes equally. Related work of multi-label few-shot learning for ACD \cite{hu2021multi} merely uses attention mechanism to denoise the sentence at word-level, but just denoising over words is not enough, since variance of noise between sentences still exists.

Moreover, in the context of few-shot text classification tasks, there
are several papers that incorporate label text information and have
obtained promising results. In the work exemplified by
\cite{zhao2022label}, although a higher boost in word-level attention
using label embedding was obtained, it still has some
deficiencies. We point out that there are many semantically similar or poorly expressed labels in the Yelp dataset we are using. For instance, the labels of \textit{food\_food\_meat} and \textit{food\_food\_chicken} are semantically similar in their
label texts, which may lead to confusion in the
classification. Furthermore, there are labels whose meanings are
rather obscure or ambiguous. Take the label
\textit{drinks\_alcohol\_hard} as an example, it is known that the
word “\textit{hard}” is a polysemous word. The word “\textit{hard}” in
this class name is related to hardness, difficulty, etc. It is obvious
that ``{\it hard}'' modifies ``{\it alcohol}'', but the word order is reversed, which may confuse the classier.  In this case, augmenting the label name with a word related to the label, such as ``{\it vodka}'', can give more specific meaning to the label name and assist separation from other aspects. 

For the purpose of tackling all the issues mentioned above, we propose a novel model named \textbf{S}entence-\textbf{L}evel \textbf{W}eighted prototypical network with \textbf{L}abel \textbf{A}ugmentation (Proto-SLWLA) that can well solve the current multi-label few-shot ACD task. Our model mainly consists of two parts, LA and SLW. Specifically, in the LA part, we concatenate the synonyms obtained from the original label text with the label itself and incorporate it into the existing word-level attention. The augmented label words will add certain auxiliary information to the label, which will make the label information become more adequate. For the SLW part, we propose assigning corresponding weights to different sentences in a class inspired by \cite{ji2020improved}, which likewise treats the samples in one class as differentiated individuals. After mitigating noises at word-level, we implement our idea as giving lower weights to sentences with more noise and higher weights to sentences with less noise, by means of a sentence-level attention mechanism. Then prototypes by giving weighted averages to the instances are obtained. With these two methods, the prototype can be more representative of the current class. Our experiments conducted on Yelp\_review \cite{bauman2017aspect} dataset shows that our method Proto-SLWLA outperforms the baselines in nearly all conditions, which demonstrates the effectiveness of our method.

The rest of this paper is organized as follows: Section 2 covers related work. Section 3 describes the proposed method of Proto-SLWLA.  In Section 4, performance evaluation of Proto-SLWLA and comparision with baseline methods are shown. Section 5 presents concluding remarks and future work.   

\section{Related Work}

\subsubsection{Aspect Category Detection} Previous research on ACD has concentrated on a single aspect, which includes unsupervised and supervised methods. Unsupervised methods use semantic association analysis based on pointwise mutual information \cite{su2006using} or co-occurrence frequency \cite{hai2011implicit,schouten2017supervised} to extract aspects. However, these approaches require large corpus resource and the performance is hardly satisfactory. Supervised methods exploit representation learning \cite{zhou2015representation} or topic-attention network \cite{movahedi2019aspect} to identity different aspect categories. In practice, these methods have shown to be effective, yet they heavily rely on a massive amount of labeled data for each aspect to train discriminative classifiers. In addition, a review sentence often encompasses multiple aspects due to the diversity and arbitrariness of human expression, which motivates the multi-label aspect category detection.

\subsubsection{Few-shot Learning} FSL is a paradigm to solve the problem of scarcity of data. Meta-learning for solving FSL problems has been widely adopted, notably in model-based approaches, optimization-based approaches, and metric-based approaches. In our paper, we concentrate on the metric-based method, whose representative models are matching network \cite{vinyals2016matching}, relation network \cite{sung2018learning}, prototypical network \cite{snell2017prototypical}, and so forth. An essential element of their idea is to learn a feature mapping function in which support and query samples are projected into an embedding space, and to classify queries by learning some metrics in that space.

\subsubsection{Multi-Label Few-Shot Learning} Compared to single-label FSL, the potential of multi-label FSL is yet to be stimulated. In the NLP domain, Proto-HATT \cite{gao2019hybrid} is proposed for an intent classification task, while designing a meta calibrated threshold mechanism with logits adaption and kernel regression. Proto-AWATT \cite{hu2021multi} focuses on multi-label few-shot aspect category detection and it is also the first work to focus on this task. It utilizes attention mechanisms to alleviate noise aspects,  achieving remarkable results. However, its prototypical network assigns an equal weight to all samples, even if certain samples contain abundance of noises and multiple aspects. This is relatively disadvantageous for a multi-label few-shot learning task. In our work, we introduce a sentence-level attention module to give different weights to different instances.

\subsubsection{Using Label Information for Text Classification} Label embedding is currently widely used in NLP for text classification tasks \cite{wang2018joint} to enhance generalization ability, and it is also very common in zero-shot and few-shot settings. In the context of zero-shot learning, prompt-based strategies \cite{puri2019zero}\cite{schick2020exploiting} to match text against class names in an implicit way have been developed. For few-shot learning, \cite{luo2021don} extracts semantics of class names and simply appends class names to the input support set and query set sentences to guide the feature representation. A work close to our task is \cite{zhao2022label} which takes label embeddings as a supplementary information in its LAS and LCL part. In our work, we extend its LAS part by applying a label augmentation method to expand the label information. 

\begin{figure}[!t]
	\centering
	\includegraphics[width=1\textwidth]{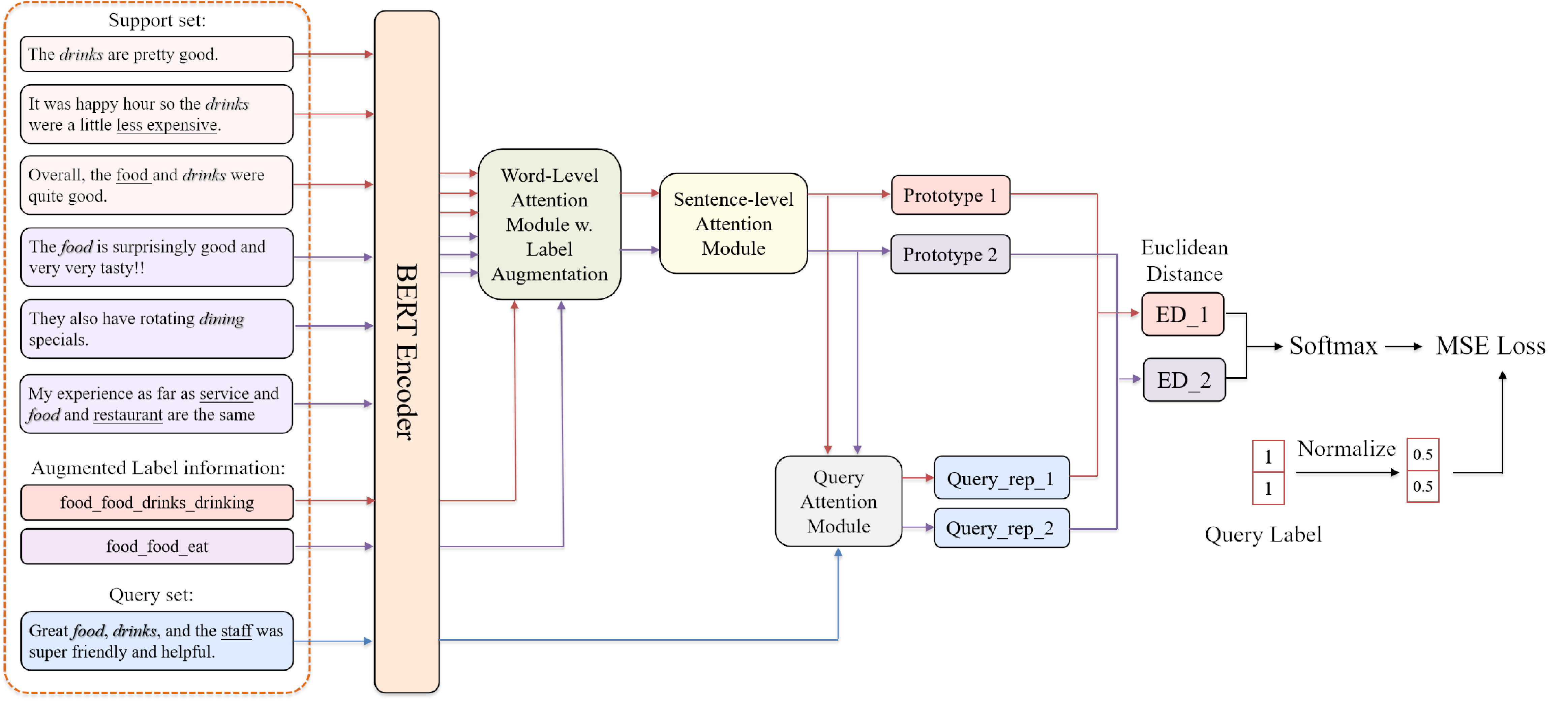}
	\caption{General architecture of our proposed model.} \label{fig1}
\end{figure}

\section{Methodology}
\subsection{Overview}
A meta-task consists of a support set and a query set. We assume that in an $ N $-way $ K $-shot meta-task, the support set is denoted as $ S = \{(x^n_1, x^n_2, ..., x^n_K), y^n\}^{N}_{n=1} $, where $ x^n_k $ represents the $ k $-th sentence in $ n $-th class and $ y^n $ is the common aspect that all $ x^n $ sentences contain. The query set is denoted as $ Q = \{(x_m, y_m)\}^M_{m=1} $, where $ x_m $ indicates a query instance and $ y_m $ is its corresponding $ N $-bit binary label from the support classes.\\

Our proposed model mainly consists of three components, which are support-set attention (word-level attention), sentence-level attention and query-set attention modules, as illustrated in Fig.\ref{fig1}(a). Given a sentence $ x = [w_1, w_2, ..., w_l] $ with length $ l $, we utilize the BERT \cite{devlin2018bert} pre-trained model as the encoder and obtain an embedding matrix $ H = [\bm{h}_1, \bm{h}_2, ..., \bm{h}_l] $, where $ H \in \mathbb{R}^{d\times L} $ and $ L $ is the maximum length of BERT input.

\begin{figure}[!t]
	\centering
	\includegraphics[width=1\textwidth]{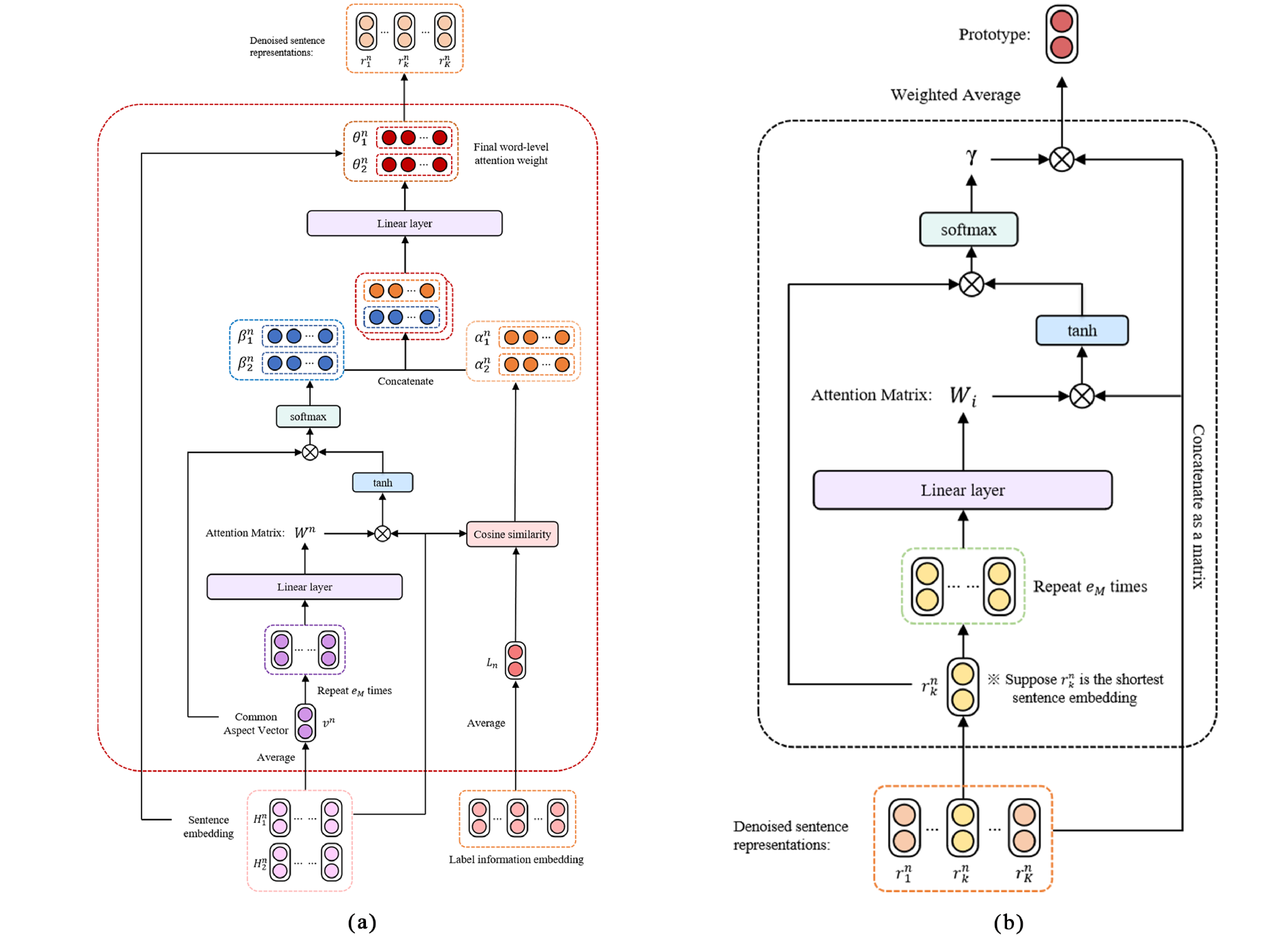}
	\caption{Specific framework of our proposed model. (a) Structure of the word-level attention module. (b) Structure of the sentence-level attention module.} \label{fig2}
\end{figure}

\subsection{Word-level Attention with Label Augmentation}
\subsubsection{Support Set Attention}
Following the work of \cite{hu2021multi}, we primarily alleviate the noises in the support set by using support set attention (word-level attention). We extract the common aspect vector $ \bm{v}^n \in \mathbb{R}^d$ out of the $ K $-shot instances by mean pooling each instance and then perform a word-level average on the $ K $ instances.

\begin{equation}
	\bm{v}^n = avg(H^n_1, H^n_2, ..., H^n_K)
\end{equation}

In order to further remove the noises, we adopt the approach of \cite{zhao2018dynamic} following \cite{hu2021multi} to train a dynamic attention matrix by feeding the repeated common aspect vector \cite{vaswani2017attention} into a linear layer. This approach is possible to learn to accommodate the common aspect and pick up on its different perspectives.

\begin{equation}
	W^n = W(\bm{v}^n \otimes e_M) + \bm{b},
\end{equation}
where $ (\bm{v}^n \otimes e_M) \in \mathbb{R}^{e_M \times d} $ denotes repeating $ \bm{v}^n $ for $ e_M $ times and $ W^n \in \mathbb{R}^{d\times d} $. The linear layer has parameter matrix $ W \in \mathbb{R}^{d \times e_M} $ and bias $ \bm{b} \in \mathbb{R}^d $. As different classes are trained, the parameters of this linear layer are constantly updated to accommodate the new classes. Then we use the common aspect vector to calculate the attention with each instance and multiply the obtained word-level weights on each sentence.

\begin{equation}\label{eq3}
	\bm{\beta}^n_k = \begin{rm}softmax\end{rm}(\bm{v}^n\tanh (W^nH^n_k)),
\end{equation}
where $ n \in [1,N] $ and $ k \in [1,K] $. So far, we have achieved a preliminary word-level attention weight to enhance the focus on the target aspect, for reducing the effect of noise aspects to some extent.

\subsubsection{Label-Guided Attention Enhanced by Label Augmentation} 
\begin{figure}[t]
	\centering\includegraphics[width=0.5\textwidth]{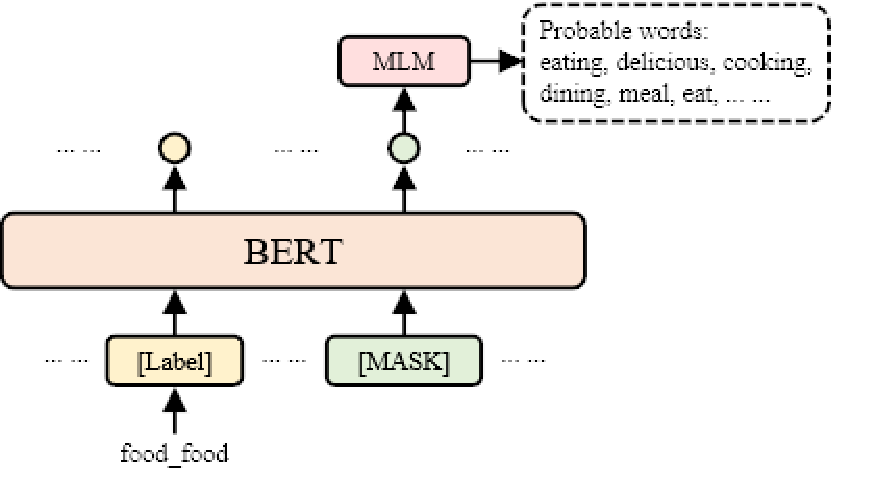}
	\caption{Using BERT masked language model (MLM) to predict words relevant to a label name.} \label{fig3}
\end{figure}

As previously stated, since there are semantically similar and ambiguous labels in the dataset, we attempt to augment label texts with supplementary words to enrich label information. In particular, in order to dig words that are relevant to the sentence as well as the label name, we design a template whose format is: “[X]. It is about [Label], and its synonym is [MASK].” In this template, [X] represents a sentence of a given aspect category in the dataset. [Label] stands for its aspect category label, and [MASK] denotes the mask token. We then supply the embedding vector $ \textit{\textbf{d}} \in \mathbb{R}^l $ into the BERT pre-trained masked language model (MLM) to predict the word that should appear at the [MASK] position, as shown in Fig. \ref{fig3}. The MLM head will output a probability distribution which indicates the likelihood of each word $w$  appearing at the [MASK] position over all the vocabulary \textit{\textbf{V}}.

\begin{equation}
	p(w|\textit{\textbf{d}}) = \begin{rm}softmax\end{rm}(W_2\sigma(W_1\textit{\textbf{d}}+\textit{\textbf{b}}')),
\end{equation}
where $ W_1 \in \mathbb{R}^{l \times l} $, $ W_2 \in \mathbb{R}^{|\textit{\textbf{V}}| \times l} $ and $ \textbf{\textit{b}}' \in \mathbb{R}^l $ are learnable parameters that have been pre-trained with the MLM objective of BERT, and $\sigma$($\cdot$) is the activation function.\\

After we have obtained a list of candidate label name related words, we filter the predicted words from each sentence by removing stop words, punctuations, words identical to the class name, etc. The final predicted augmenting words for each class name are shown in Table \ref{tab2}. Then we take the top  words of all the predicted words in each category, and find the top \textit{m} words with the highest frequency among the total number of sentences multiplied by \textit{m} words.\\

Once we obtained these words, we append them to the original label name with an underline to form a new label. For instance, take $ m=1 $ as an example, the original label name \textit{drinks\_alcohol\_hard} will be transformed into \textit{drinks\_alcohol\_hard\_vodka}, which nicely emphasizes the meaning of “textit{liquor}”, thus eliminating its interference with other meanings in the synonym.\\

\begin{table}[t]
	\caption{Part of the finally  obtained words relevant to a label name}\label{tab2}
	\centering\resizebox{\linewidth}{!}{
		\begin{tabular}{|c|c|}
			\hline \textbf{Label Name} & \textbf{Predicted Words}\\
			\hline food\_food & eat, delicious, dining, cooking, meal, eating, foods, …\\
			\hline parking & cars, space, traffic, parking, cars, driving, bike, …\\
			\hline restaurant\_entertainment\_music & song, pop, jazz, opera, melody, rock, blues, folk, …\\
			\hline drinks\_alcochol\_hard & vodka, tequila, rum, gin, bitter, bourbon, …\\
			\hline restaurant\_location & place, destination, locality, spot, geography, …\\
			\hline entertainment\_casino & gambling, vegas, gaming, poker, casinos, game, …\\
			\hline building\_hall & hallway, lobby, corridor, wall, halls, library, …\\
			\hline
	\end{tabular}}
\end{table}

 So far, we have accomplished the process of label augmentation. Now we need to integrate augmented label information into the word-level attention to give more guidance to support sentences on label information as \cite{zhao2022label}. Specifically, we enter the augmented label information into the BERT model to obtain its embedding, and compute the cosine similarity between the label information embedding and sentence embedding:
 
 \begin{equation}
 	\bm{\alpha}^n_k = \begin{rm}cos\end{rm}(\bm{L}_n, H_k^n),
 \end{equation}
where $ \bm{L}_n \in \mathbb{R}^d $ is the label information embedding of class \textit{n} in the support set, which is calculated by averaging in terms of the length of label information. $ H_k^n $ is the word embedding matrix of the $ k $-th sentence in the $ n $-th class. We then combine the calculated cosine similarity $ \bm{\alpha}_k^n$ with the previously obtained word-level attention $ \bm{\beta}_k^n $ in (\ref{eq3}). We concatenate the two vectors and enter them into the linear layer to obtain the final attention weight.

\begin{equation}
	\bm{\theta}_k^n = W_g[\bm{\alpha}_k^n ; \bm{\beta}_k^n] + \bm{b}_g,
\end{equation}
where $ \bm{\theta}_k^n \in \mathbb{R}^l $, $ W_g $ and $ b_g $ are trainable parameters of the linear layer. [$\cdot$ ; $\cdot$] denotes the concatenation operation. We then renormalize the attention score by the softmax function to make the weight more reliable.

\begin{equation}
	\bm{\tilde{\theta}}_k^n = \begin{rm}softmax\end{rm}(\bm{\theta}_k^n)
\end{equation}

Eventually, we assign the final word-level attention weight to each sentence in the support set.

\begin{equation}\label{eq8}
	\bm{r}_k^n = \bm{\tilde{\theta}}_k^n H_k^n,
\end{equation}
where $ n \in [1, N] $ and $ k \in [1, K] $. So far, we have constructed a collection of $ R_n = [\bm{r}^n_1, \bm{r}^n_2, ..., \bm{r}^n_K] $ which consists of denoised support set representations. The whole process of this word-level attention module is illustrated in Fig \ref{fig2} (a).

\begin{figure}[t]
	\centering\includegraphics[width=0.5\textwidth]{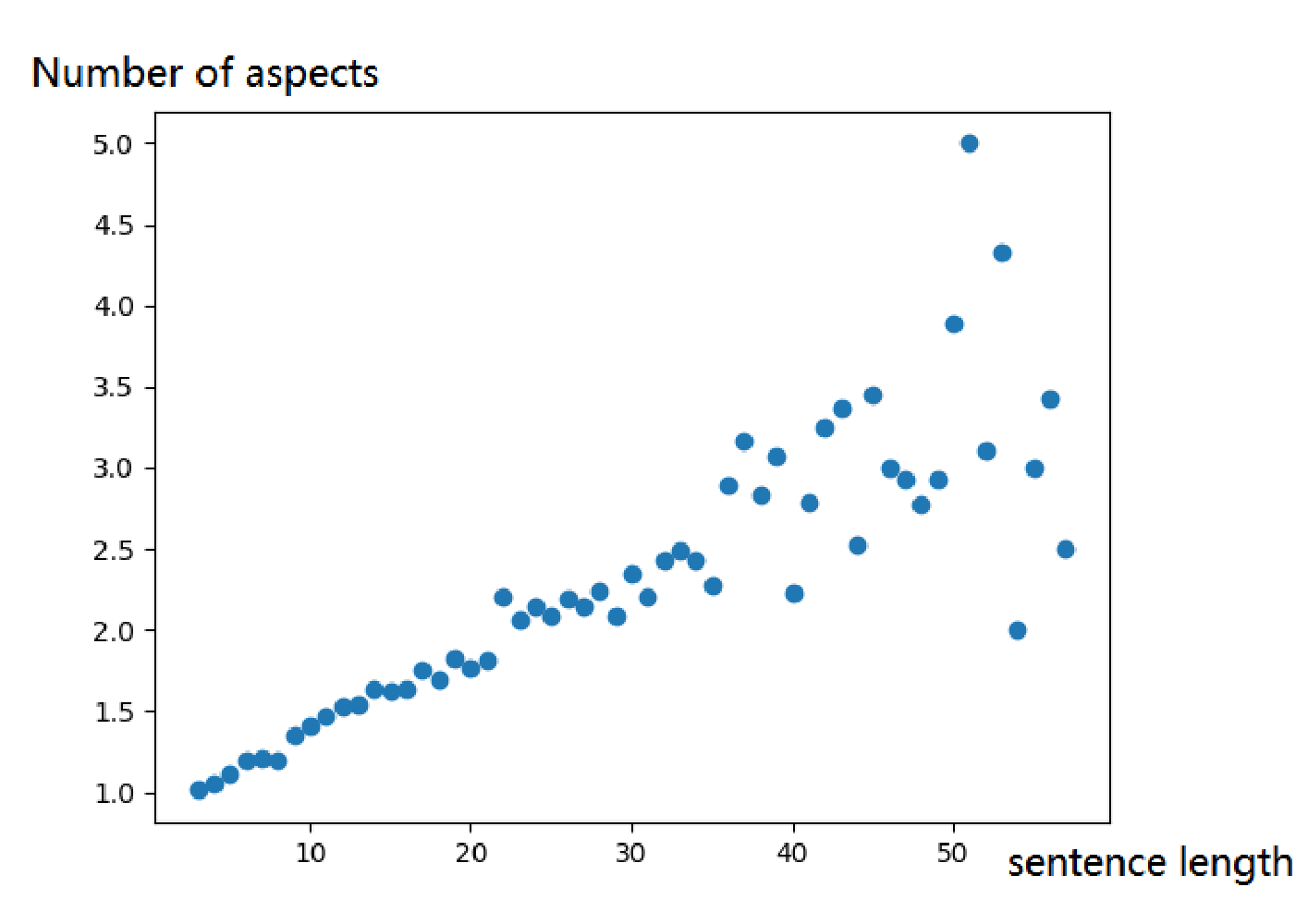}
	\caption{Distribution of sentence length and number of aspects in the Yelp  dataset. We randomly selected 8000 samples in the dataset and averaged the number of aspects for all sentences of the same length.} \label{fig4}
\end{figure}

\subsection{Sentence-level Attention}

In this section, we describe calculation of sentence-level weights. The architecture is depicted in Fig.\ref{fig2}(b). As previously mentioned, we would like to adjust weights on different sentences in the classes depending on the amount of estimated noises involved in the sentence. Hence, we introduce a method to compute the attention by centering on the shortest sentence in the class. As illustrated in Fig. \ref{fig4}, we can observe that as the length of the sentence increases, the number of aspects therein increases as well. That is, the noises of the sentence are also extending. Consequently, it is reasonable to surmise that the shorter the sentence, the fewer number of aspects the sentence contains, which will then have a larger possibility of having a single aspect. Since each sentence always contains the target aspect, the chance of having noisy aspects becomes smaller than longer sentences. \\

Thus, we introduce a mechanism that emphasizes aspects learned from the shortest sentence and apply this aspect weighting to all the sentences in the support set. We first locate the denoised support representation $ \bm{r}_{min}^n $ of the shortest sentence and then repeating it for $ e_M $ times, which aims to learn different perspectives of the shortest sentence. Then we feed it to a linear layer to obtain the attention matrix $ W_s^{n} $, where $ W_s^{n} \in \mathbb{R}^{d\times d}$ and $ W_s $, and $ \bm{b}_s $ are trainable parameters.

\begin{equation}
	W_s^{n} = W_s({\bm{r}_{min}^n \otimes e_M}) + \bm{b}_s
\end{equation}

Similarly, we follow the preceding method to directly use the shortest sentence embedding $ \bm{r}_{min}^n $ to compute sentence-level attention on all the denoised sentence representations. Particularly, $ R^n $ is multiplied with the attention matrix $ W'^n $ to exploit the relationships between the shortest sentence and other sentences from different perspectives. The weight is calculated as follows:

\begin{equation}
	\bm {\gamma}^n = \begin{rm}softmax\end{rm}(\bm{r}_{min}^n\tanh(W_s^nR^n)),
\end{equation}
where $ \bm {\gamma}^n \in \mathbb{R}^k,$ and $ R^n \in \mathbb{R}^{d\times k} $ represents the concatenation of all the denoised representations in one class as shown in (\ref{eq8}). By doing this, longer sentences which are dissimilar to the shortest sentence and meanwhile contain more noise will obtain lower weights. Finally, we perform a weighted average to the representations to derive the final prototype $ \bm{p}^n \in \mathbb{R}^d $.

\begin{equation}
	\bm{p}^n = \bm {\gamma}^nR^n
\end{equation}

In this way, we obtain $N$ prototypes $ [\bm{p}^1, \bm{p}^2, ..., \bm{p}^N] $ after processing all the classes in the support set.

\subsection{Query Attention}
Following the denoising operation of the support set, we ought to mitigate the noises in the query set as well, since the query set also contains some irrelevant aspects. To achieve this goal, we use the prototype $ \bm{p}^n $ we just obtained to compute the attention with the embedding of a query sentence $ H_m \in \mathbb{R}^{d\times L} $ and acquire the query representation $ \bm{r}_m \in \mathbb{R}^d $. What we want to accomplish is to enable query representation to be more attentive to the prototype aspect.

\begin{equation}
	\bm{r}_m = \begin{rm}softmax\end{rm}(\bm{p}^n\tanh (H_m))
\end{equation}

Up to this point, we have completed introducing all the modules of the model and finished construction of representations of a given support set and query set.

\subsection{Training Objective}
In this paper, we use the Euclidean Distance (ED) to measure the distances between prototypes and query representations. The final prediction of a query instance is the negative distances and then we use a softmax to normalize the result.

\begin{equation}
	\hat{\bm{y}} = \begin{rm}softmax\end{rm}(-\begin{rm}ED\end{rm}(\bm{p}^n, \bm{r}_m)),
\end{equation}
where $ n \in [0, N] $, and $ m \in [0, M] $. Lastly, we use the mean square error (MSE) loss to be our final training objective.

\begin{equation}
	L = \sum (\bm{\hat{y}} - \bm{y}_m)^2,
\end{equation}
where $ \bm{y}_m $ is the $ N $-bit golden label for query instance $ x_m $. Note that since $ \bm{\hat{y}} $ is softmaxed, our golden label $ \bm{y}_m $ should be normalized as well. During the training process we allow the predicted values to be as close as possible to the golden label.

\section{Experiments}
In this section, we primarily introduce the dataset used in our work, together with baselines, evaluation metrics, and implement details. Thereafter, we present and analyze the experimental results on our dataset in four different settings.

\subsection{Dataset}
Since the Yelp dataset having review aspects and used in \cite{hu2021multi} is not publicly available, we construct the dataset by combining the Yelp\_dataset\_round8 and Yelp\_review\_aspect \cite{bauman2017aspect} which are datasets consisting of extensive user reviews. After processing the raw data into sentences and their corresponding aspects, we collected the sentences for each aspect and selected 100 aspects from all the 135 aspects and remove 35 of them. The selected aspects are split without intersection into 64 aspects for training, 16 aspects for validation, and 20 aspects for testing. We randomly sample 800 meta-tasks from the 64 gathered aspect sentences for training, 600 meta-tasks from the 16 gathered aspects for validation and 600 meta-tasks from the 20 gathered aspects for testing, following \cite{hu2021multi}. The statistics of our dataset is shown in Table \ref{tab3}. Note that at each epoch of training, the 800 meta-tasks are resampled.

\begin{table}[t]
	\caption{Statistics of the Yelp dataset. \textbf{\#cls.} indicates the number of classes. \textbf{\#inst./cls.} indicates the number of instances per class. \textbf{\#inst.} indicates the total number of instances.}\label{tab3}
	
	\centering\resizebox{0.4\linewidth}{!}{
		\begin{tabular}{|c|c c c|}
			\hline \textbf{Dataset} & \textbf{\#cls.} & \textbf{\#inst./cls.} & \textbf{\#inst.}\\
			\cline{1-4} FewAsp & 100 & 630 & 63000\\
			\hline
	\end{tabular}}
\end{table}

\subsection{Baseline Models}
Our method is compared with the following methods: Matching Network \cite{vinyals2016matching}, Relation Network \cite{sung2018learning}, Prototypical Network \cite{snell2017prototypical} and Proto-AWATT w/o DT \cite{hu2021multi} and Proto-SLW. Note that we use BERT as the encoder for all baseline models for the sake of fairness.

\subsubsection{Matching Network \cite{vinyals2016matching}}
It learns an embedding mapping function first, combines the samples of support set and query set samples and enters them into Bi-LSTM, and finally adopts the cosine similarity as the distance measure to obtain the classification result.

\subsubsection{Relation Network \cite{sung2018learning}}
Instead of a fixed distance metric, it uses a deep neural network with multiple layers of convolution to compute the relationship between query samples and support samples.

\subsubsection{Prototypical Network \cite{snell2017prototypical}}
By averaging the corresponding support samples, it computes a prototype for each class and uses the negative Euclidean distance between the query samples and the prototype for the few-shot classification task.

\subsubsection{Proto-AWATT \cite{hu2021multi}}
It is the first approach for multi-label aspect category detection tasks. It mitigates the adverse effects caused by noisy aspects using support set and query set attention mechanisms.

\subsubsection{Proto-SLW}
As an ablation setting, this model is removed of the LA part from our proposing model and only utilizing the sentence-level attention to assign different weights to different sentences in one class in the support set.

\subsubsection{Proto-SLW+LAS}
We add the LAS part from \cite{zhao2022label} to our SLW model to take the label name itself as a complementary information of the attention weights in the support set. Note that this case is actually equivalent to the case of \textit{m}=0 in Proto-SLWLA.

\subsection{Evaluation Metrics}
Traditional single-label FSL tasks in the past typically used accuracy to measure the performance of the model. In the multi-label task, we follow Proto-AWATT and choose the AUC (Area Under Curve) score which is used to select model and macro-F1 score as evaluation metrics.

\subsection{Experimental Settings}
For parameter settings, we set $ m = 1 $, $ d = 768 $, $ L = 50 $, $ e_M = 4 $, $ Q = 5\times N $. We train our model on GeForce RTX 3090 GPU and set our learning rate as 1e-5, batch size as 4 when $ N = 5 $, as 2 when $ N = 10 $. When performing label augmentation, we randomly select 2000 sentences in the query set of each class for subsequent operations.\\

Regarding the threshold value, we set $ \tau = 0.3 $ in all the conditions. We adopt an early stop strategy when the AUC score is no longer increased in 3 epochs. Then we will select the epoch which has the best result of the AUC score in the validation phase for testing.

\begin{table}[t]
	\caption{Experimental results of our model with AUC and macro-F1(\%) evaluated on FewAsp. m represents for the number of words augmented by each label. The symbol $ \dag $ indicates  $ p $-value<0.05 of the T-test comparing with Proto-AWATT, while symbol $ \ddag $ indicates that $ p $-value<0.05 of the T-test comparing with Proto-SLW.}\label{tab4}
	
	\resizebox{\linewidth}{!}{
		\begin{tabular}{|c|c|c|c|c|c|c|c|c|}
			\hline \multirow{2}{*}{ Models } & \multicolumn{2}{c|}{ 5-way 5-shot } & \multicolumn{2}{c|}{ 5-way 10-shot } & \multicolumn{2}{c|}{ 10-way 5-shot } & \multicolumn{2}{c|}{ 10-way 10-shot } \\
			\cline { 2 - 9 } & AUC & F1 & AUC & F1 & AUC & F1 & AUC & F1 \\
			\hline Matching Network & $0.9025$ & $66.59$ & $0.9230$ & $70.97$ & $0.8834$ & $51.54$ & $0.9085$ & $53.84$ \\
			\hline Prototypical Network & $0.9017$ & $65.71$ & $0.9318$ & $71.55$ & $0.8991$ & $53.82$ & $0.9063$ & $55.44$ \\
			\hline Relation Network & $0.8463$ & $54.72$ & $0.8473$ & $55.54$ & $0.8428$ & $42.92$ & $0.8325$ & $45.86$ \\
			\hline Proto-AWATT w/o DT & $0.9061$ & $66.32$ & $0.9319$ & $71.67$ & $0.8999$ & $53.86$ & $0.9125$ & $57.75$ \\
			\hline Proto-SLW & $0.9116$ & $67.27$ & $0.9387$ & $72.83$ & $\mathbf{^{\textcolor{white}{\dag}}0.9062^{\textcolor{white}{\dag}}}$ & $54.74$ & $0.9156$ & $57.33$ \\
			\hline Proto-SLW+LAS ($ m=0 $) & $ 0.9123 $	& $ 67.94 $ & $ 0.9374 $ & $ 72.90 $ & $ 0.9037 $ & $ 54.98 $ & $ 0.9119 $ & $ 57.09 $ \\
			\hline Proto-SLWLA ($ m=1 $) & $ 0.9156 $ & $ 68.11 $ & $ \bm{0.9391^{\dag}} $ & $ \bm{73.20^{{\dag}{\ddag}}} $ & $ 0.9024 $ & $ 54.79 $ & $ \bm{0.9179^{{\dag}{\ddag}}} $ & $ \bm{58.69^{{\dag}{\ddag}}} $ \\
			\hline Proto-SLWLA ($m=2$) & $ \bm{0.9157^{{\dag}{\ddag}}} $ & $ \bm{68.30^{{\dag}{\ddag}}} $ & $ 0.9377 $ & $ 72.87 $ & $ 0.9026 $ & $ 55.18 $ & $ 0.9156 $ & $ 58.20 $ \\
			\hline Proto-SLWLA ($m=3$) & $ 0.9117 $ & $ 67.57 $ & $ 0.9380 $ & $ 72.97 $ & $ 0.9038 $ & $ \bm{55.63^{{\dag}{\ddag}}} $ & $ 0.9154 $ & $ 57.94 $ \\
			\hline
		\end{tabular}}
\end{table}

\subsection{Experimental Results and Discussions}
The experimental results are shown in Table \ref{tab4}. The results demonstrates that both our word-level attention with label augmentation module and sentence-level attention module are effective. It is worth mentioning that in the 10-way 10-shot scenario, Proto-SLWLA(m=1) improves  F1 score by 1.36\% from Proto-SLW, which is considered a significant improvement in our experimental results. \\

When Proto-SLW is compared with Proto-AWATT, the results of Proto-SLW are better than those of Proto-AWATT in all four scenarios, which fully illustrates the effectiveness of our sentence-level attention module. Furthermore, the result on 10-shot has slightly more improvement than that of the 5-shot result. This suggests that a noise-less short sentence is more likely to be included in more shots (sentences). However, the 10-way boost is a little less pronounced than the 5-way, since 10-way (10-classes) causes more chance for short sentences to contain noise aspects than 5-way (5-classes). \\

Compared with Proto-SLW+LAS, the results of Proto-SLWLA outperform in all these four scenarios. This indicates that our LA part is effective, and suggests that it is reliable to use label-related words to enhance the label itself. In addition, for Proto-SLWLA we evaluated with three different values of $m$, which represents the number of words augmented by each label. We can observe that our model can achieve the best results when $ m=1 $ or $ 2 $, whereas for $ m=3 $, the performance of the model is decreasing and sometimes even lower than Proto-SLW+LAS. This demonstrates that the first or the second of the augmenting words are highly relevant to the label. However as the number of augmenting words increases, lower-ranked words seem to be causing drifts on label semantics. \\

\section{Conclusion}
In this paper, we proposed Proto-SLWLA, which is based on prototypical network with sentence-level weighting and label augmentation to tackle the multi-label few-shot aspect category detection task. Existing methods in this domain often utilize prototypical network, but they perform denoising merely at the word-level and do not focus on the variations between instances.  Since the concentrations of noises in the target aspects are varying between instances, we introduced sentence-level attention to assign specific weights to the instances after using word-level attention.  Also, another existing approach incorporates label text information into the word-level attention module to improve the performance, but the label name texts are not sufficient, because label names are often semantically similar and ambiguous each other, causing separation hard in the representation space.  To improve separation by label names, we introduced label name augmentation in which  a template is designed and masked language model prediction is utilized to generate words related to each label name, and append them to the label information, which is then used for a guidance of the word-level weights. Our experimental evaluations by AUC and macro-F1 score demonstrate that our design is feasible and effective, outperforming nearly all the baseline models. 
%
%
%
%
\clearpage
\bibliography{ref}
\end{document}